\documentclass[runningheads]{llncs}
\usepackage[T1]{fontenc}
\usepackage{graphicx}
\usepackage{subcaption}

\usepackage{graphicx}
\usepackage{amsmath}

\begin{document}
\title{Quasi-static analysis of passive stability in a novel underactuated multi-finger hand}
\titlerunning{Quasi-static analysis of a novel underactuated multi-finger hand}
%
\author{Léonie Plancoulaine\inst{1}\orcidID{0009-0001-9540-4141} \and \\ 
Sylvain Guegan \inst{2}\orcidID{0000-0002-8797-6078} \and
Franck Plestan \inst{1}\orcidID{0000-0001-8971-5106} \and \\ Damien Chablat \inst{1}\orcidID{0000-0001-7847-6162}}
\authorrunning{L. Plancoulaine et al.}
%
\institute{Nantes Université, Ecole Centrale de Nantes, CNRS, LS2N, UMR  6004, Nantes, France \email{leonie.plancoulaine@ls2n.fr, damien.chablat@cnrs.fr, franck.plestan@ls2n.fr}\\
\and
Université Rennes, INSA Rennes, LGCGM, F-35000 Rennes, France
\email{sylvain.guegan@insa-rennes.fr}}
\maketitle              
\begin{abstract}
Underactuated robotic hands achieve adaptive and robust grasping with a reduced number of actuators, but predicting the stable equilibrium pose of the grasped object remains a significant challenge. This paper introduces a quasi-static analytical approach to assess passive stability in underactuated multi-finger hands. {A novel three-finger hand architecture integrating a differential spring-loaded slider mechanism is introduced, enabling versatile and adaptive grasping. The study focuses on how the differential mechanism influences the overall grasp behavior and analyzes the effect of object size on the stable equilibrium configurations for two canonical grasp types: cylindrical and spherical.} 

\end{abstract}
%
%
%
\section{Introduction}
\label{sec:1}
The human hand remains one of the most challenging parts of the human body to reproduce robotically due to its remarkable dexterity. It can perform up to 34 complex grasp types \cite{Feix}, which depend on both the object geometry and the intended task. These grasps fall into two categories: power grasps, which generate large forces through multiple contacts across the phalanges and the palm, and precision grasps, which rely on the distal phalanges for fine control. Moreover, the relative orientation of the fingers defines different canonical grasp geometries: cylindrical grasps occur when two fingers oppose a third, while spherical grasps arise when all three fingers converge toward a common point \cite{Miller}.
A growing design philosophy \cite{Birglen} aims to achieve human-like dexterity with the simplest possible architecture. Underactuated hands \cite{Gosselin} represent a compromise solution, using fewer actuators than degrees of freedom (DOF) and relying on passive elements, such as springs, to control the remaining joints. This approach simplifies the mechanical design, reduces cost and weight, and enables natural shape adaptation, though often at the expense of precision. 
{This paper presents the design of a new three-finger underactuated hand with two-phalanx fingers capable of passive spatial motion. In the proposed hand, a differential spring-loaded slider \cite{Massa} was chosen to distribute a single input actuation force among the fingers. While several alternative mechanisms exist \cite[Chap.~6]{Birglen}, including pulleys, bar structures, and gear systems, the spring-loaded slider is an efficient solution, enabling all three outputs with minimal weight.}
Stability analysis \cite{Roa} is a key tool for identifying suitable grasp configurations for a given object, as well as for evaluating grasp planning strategies. Analyzing equilibrium configurations provides insight into object size limitations and into the influence of mechanical design choices on grasp performance. {Although differential mechanisms are widely used, their impact on grasp stability remains largely unexplored. Yet, this influence is central to understanding how forces are redistributed during contact.}

{This paper investigates the passive stability of the proposed underactuated hand for two canonical grasp types: cylindrical and spherical. The analysis focuses on the influence of the differential mechanism on the overall grasp behavior, the object’s stable equilibrium positions, and the range of admissible object sizes.}
\section{Hand architecture}
\begin{figure}[!ht]
    \includegraphics[width=1\linewidth]{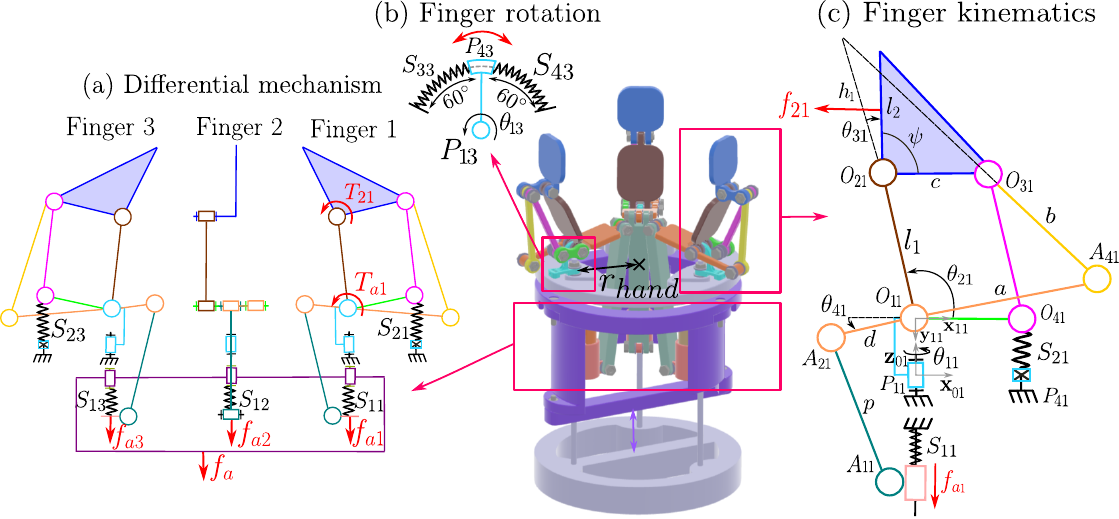}
    \caption{{Overview of the hand kinematics and the three levels of underactuation: (a) the differential mechanism, (b) the finger rotation, and (c) the finger kinematics.}}
    \label{fig:finger}
\end{figure}

This section presents the hand{\footnote[1]{Animation of the different levels of underactuation: https://youtu.be/29pfHB39d48}} developed by the authors (Fig.~\ref{fig:finger}), composed of three fingers arranged on a circle of radius $r_{hand} = 50 \text{mm}$, spaced $120^\circ$ apart. {The underactuation occurs at three levels: between fingers, passive rotation of fingers, and underactuated fingers. }
{Firstly, underactuation between fingers is achieved thanks to a differential spring-loaded slider, following the approach proposed by Massa \cite{Massa} as indicated above.} It includes a sliding element that is actuated manually and later will be driven by an electric motor, and springs {$S_{1i}$}, {where $i$ denotes the $i^\text{th}$ finger}. The actuation of the hand controls the opening/closing of the fingers. If the motion of one finger is blocked, the corresponding spring compensates by absorbing the displacement, thereby allowing the other two fingers to continue moving.  Unlike the slider proposed by Massa \cite{Massa}, the one presented here is connected to mechanical linkages, enabling torque transmission thanks to a lever-like motion. {The parameters $d = 37.8\,\text{mm}$ and $p=101\,\text{mm}$ are the lengths $[A_{2i}O_{1i}]$, and $[A_{1i}A_{2i}]$, respectively, and $\theta_{4i}$ is the angle between $\mathbf{x}_{1i}$ and $[A_{2i}O_{1i}]$.}

{Secondly, most underactuated fingers operate in a planar configuration, which limits grasp diversity \cite{Piazza}. However, some designs incorporate additional degrees of freedom, such as controlled rotations of the fingers \cite{Laliberte} or passive abduction/adduction \cite{Hamon}. While these enhancements increase grasp diversity, they often introduce additional kinematic complexity.} The authors add a revolute joint passively actuated {by antagonistic} springs {$S_{3i}$ and $S_{4i}$}, at the base of each finger (point $P_{1i}$). This joint enables three-dimensional motion without significantly increasing mechanical complexity.

{Finally, the finger kinematics, based on the theory of Birglen \cite[Chap.~4]{Birglen}, enable both precision and power grasps. In the last case, spring $S_{2i}$ plays a key role: its extension drives the movement of the distal phalanx.} The finger’s structure is composed of two interlinked closed loops: the actuation loop ($O_{1i}$, $A_{4i}$, $O_{3i}$, $O_{2i}$) drives the motion, while the parallel grasp loop ($O_{1i}$, $O_{4i}$, $O_{3i}$, $O_{2i}$) enables both precision and power grasps. This study focuses specifically on precision grasp. 
The lengths $l_1 = 50.7 \text{mm}$ and $l_2=38 \text{mm}$ correspond to the lengths of the proximal and distal phalanges, respectively, while $\psi = 90^\circ$ defines the constant angle between the distal phalanx and $[O_{2i}O_{3i}]$. The parameters $a = 33.5 \text{mm}$, $b=54 \text{mm}$, $c = 15\text{mm}$ are the lengths of $[O_{1i}A_{4i}]$, $[A_{4i}O_{3i}]$, $[O_{2i}O_{3i}]$, respectively. 

The finger possesses three DOF, defined as follows: $\theta_{1i}$, the angle between axis ${\bf x}_{0i}$ and ${[P_{1i}P_{4i}]}$ along ${\bf z}_{0i}$; $\theta_{2i}$, the angle between axis ${\bf x}_{1i}$ and ${(O_{1i}O_{2i})}$ along ${\bf z}_{1i}$; $\theta_{3i}$, the angle between the proximal and distal phalanges. For a precision grasp $\theta_{3i}=-\theta_{2i}-\psi$ { as the distal phalanx remains vertical.}
\section{Method}
\label{sec:method}
\begin{figure}
    \includegraphics[width=0.9\linewidth]{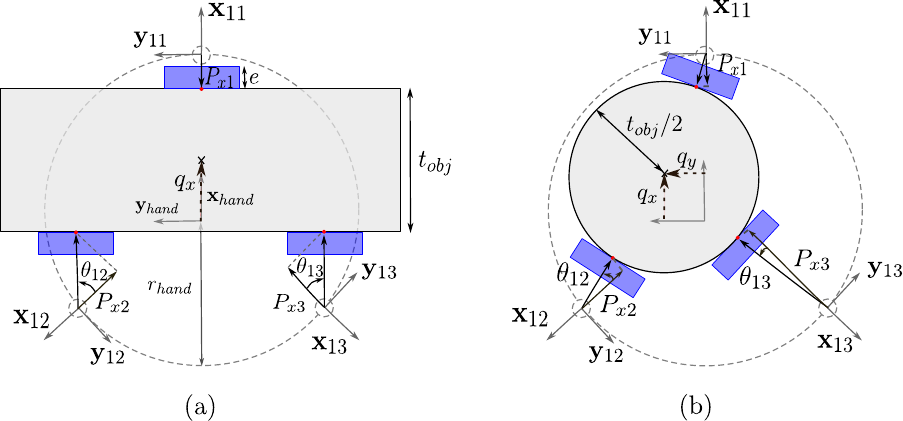}
    \centering
    \caption{{The two grasping scenarios: (a) cylindrical grasp, and (b) spherical grasp.}}
    \label{fig:methode}
\end{figure}
The grasping performance analysis is conducted in 2D, focusing on the object’s stable position with respect to the hand frame $\mathbf{x}_{hand}$, $\mathbf{y}_{hand}$ (Fig.~\ref{fig:methode}). 
Although friction at the fingertip-object and object-table interfaces may influence grasp stability, it is neglected in the present study to focus on kinematic effects and the role of the passive differential mechanism.  
{Nevertheless, it is useful to briefly discuss their effects during the grasping process. While the object remains on the table, object-table friction may contribute to stabilizing its position during finger closure. Once the object is lifted off the surface, this friction force vanishes, and the grasp relies solely on the contact forces and friction at the fingertip-object interfaces. As a result, the object may slightly reposition within the hand until reaching a new equilibrium position. Furthermore, fingertip-object friction may prevent slipping and helps maintaining a stable grasp}. Moreover, the contact between the object and the finger is assumed to occur at a single point located at the midpoint of the distal phalanx. {This assumption corresponds to a worst-case scenario as it reduces the range of stable grasps.} Then, the distance between the revolute joint at $O_{2i}$ and the contact point is $k_{2i}=l_2/2$. Furthermore, $T_{2i}$, the torque induced by the spring {$S_{2i}$, is neglected because the spring remains nearly unstretched, generating little to no torque.} 
The goal is then to determine the contact force generated by each finger. According to Birglen's results \cite[Chap.~4, Eq.~(5)]{Birglen}, the finger-object contact forces are given by:
\begin{equation}
    f_{2i}=\frac{h_iT_{ai} }{(h_i + l_{1}) k_{2i}}.
\label{f2i}
\end{equation}

{Firstly, the finger actuation torque, $T_{ai}$, which arises from the differential mechanism and the lever-arm effect, is defined as $T_{ai}=f_{ai}d\cos\theta_{4i}$.} $f_{ai}$, the finger actuation force, depends on the spring forces $f_{si}$ generated by the springs $S_{1i}${, each with a stiffness of K =5 N/mm}, as well as on the global actuation force $f_a = 50 \text{N}$. According to Birglen \cite{Birglen}, the three output forces can be expressed as:
\begin{equation}
\begin{bmatrix}
f_{a1} \\
f_{a2} \\
f_{a3}
\end{bmatrix}
=
\begin{bmatrix}
1 & 0 & -1&-1 \\
1 &  -1 &  0&-1 \\
1 &  -1 & - 1&0
\end{bmatrix}
\begin{bmatrix}
f_{a} \\
f_{s1}\\
f_{s2} \\
f_{s3}
\end{bmatrix}.
\label{eq:distrib}
\end{equation}
From the geometry of the mechanism, one can derive the expression of $\theta_{4i}$ as:
\begin{equation}
    \theta_{4i}=\arccos\left(\frac{g_i^2+c^2-l_1^2}{2g_ic}\right)-\arccos\left(\frac{g_i^2+a^2-b^2}{2g_ia}\right)
\label{theta4}
\end{equation}
with $g_i=\sqrt{(c-l_1\cos\theta_{2i})^2+(l_1\sin\theta_{2i})^2}$.
Secondly, $h_i$ denotes the signed distance between $O_{2i}$ and the intersection of the lines $[O_{1i}O_{2i})$ and $[A_{4i}O_{3i})$ (Fig.~\ref{fig:finger}). {As detailed by Birglen \cite[Chap.~4, Eq.~(6)]{Birglen}, $h_i$ is a function of the finger configuration and in particular on the joint angle $\theta_{2i}$.} The proximal phalanx flexion can be determined in both scenarios from the geometry:
\begin{equation}
    \theta_{2i}=-\arccos\left(\frac{P_{xi}}{l_1}-\frac{\cos\psi}{k_{2i}l_1}\right)
\label{theta2}
\end{equation}
$P_{xi}$ is the relative distance between the distal phalanx and the axis $\mathbf{z}_{0i}$. For each scenario, this distance can be determined {from the geometric configuration shown in Fig.~\ref{fig:methode}.  It depends on the object width $t_{obj}$, the hand radius $r_\text{hand}$, the finger thickness $e= 5.5 \text{mm}$, and the object's stable position along $\mathbf{x}_\text{hand}$ relative to the hand center, $q_x$:}
\begin{equation}
\begin{aligned}
P_{x1} = \frac{-r_{hand} + q_x}{\cos\theta_{11}} + \frac{t_{obj}}{2} + e;\qquad
P_{x2,3}=\frac{-r_{hand}/2-q_x}{\cos\alpha_i}+\frac{t_{obj}}{2}+e.
\end{aligned}
\label{pointcontcyl}
\end{equation}
For finger 2 and finger 3, one can define: $\alpha_2=\frac{\pi}{3}-\theta_{12}$ and $\alpha_3=\frac{\pi}{3}+\theta_{13}$.

\textbullet\ \textbf{Cylindrical grasp:}  
In this scenario, the object to be grasped is modeled as a rectangular prism. It is then assumed that finger~1 is positioned in front of fingers~2 and~3, leading to the angular configuration $\theta_{11} = 0^\circ$, $\theta_{12} = 60^\circ$, and $\theta_{13} = -60^\circ$.

\textbullet\ \textbf{Spherical grasp:}  
The object under study is a cylinder. 
{Similarly to $q_x$, $q_y$ is the object position of stability along $\mathbf{y}_\text{hand}$ relative to the hand center. The angular position can be expressed according to Fig.~\ref{fig:methode}:}
\begin{equation}
\begin{aligned}
\theta_{11} = \arctan\left(\frac{q_y}{r_{hand}-q_x}\right), \quad
\theta_{12}= \frac{\pi}{3}- \arctan\!\left(\frac{\sqrt{3}r_{\text{hand}} - 2q_y}{r_{\text{hand}} + 2q_x}\right), \\
\theta_{13}= -\frac{\pi}{3} + \arctan\!\left( \frac{\sqrt{3}r_{\text{hand}} + 2q_y}{r_{\text{hand}} + 2q_x}\right).
\end{aligned}
\label{point}
\end{equation}

The objective is then to resolve the static equation for each grasp configuration:  $\mathbf{f}_{21}+\mathbf{f}_{22}+\mathbf{f}_{23}=\mathbf{0}$. 
Thanks to these, the stable object position and the finger configuration can be determined.
For cylindrical grasp, it is assumed that fingers~2 and~3 exert the same force, $\mathbf{f}_{22} = \mathbf{f}_{23}$, which implies $\mathbf{f}_{s2} = \mathbf{f}_{s3}$. 
Furthermore, in order to assess the stability of the equilibrium configuration in both grasp scenarios, and assuming the local behaviour around the equilibrium is conservative, a local Hessian of the potential energy {\cite{Kragten}} is examined.
\section{Results}
\begin{figure}[htbp]
\centering
\begin{subfigure}{0.23\textwidth}
    \centering
    \includegraphics[width=\linewidth]{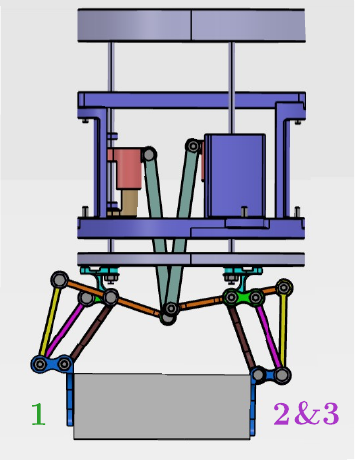}
\end{subfigure}
\hfill
\begin{subfigure}{0.52\textwidth}
    \centering
    \includegraphics[width=\linewidth]{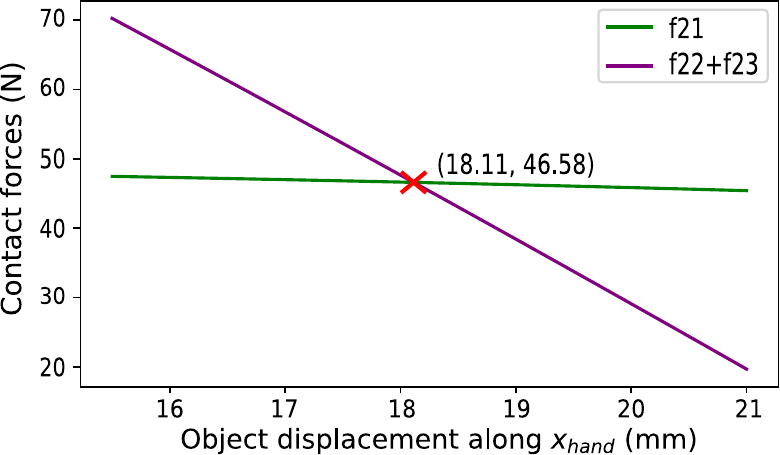}
\end{subfigure}
\caption{Contact forces as a function of the object displacement along $x_{hand}$ for a prism of width 110 mm.}
\label{fig:twofigures}
\end{figure}
\noindent\textbf{Cylindrical grasp:}
{Firstly, Fig.~ \ref{fig:twofigures} shows the contact forces during the grasping of a rectangular prism with a width of $t_{obj}=110 \text{mm}$ as a function of the object displacement. It can be observed that a stable position is reached for a displacement $q_x$ of $18.11 \text{mm}$. This offset results from the finger arrangement on the hand, the finger kinematics and the action of the differential mechanism with the compression of the spring $S_{1i}$. This compression occurs because finger~1 opposes fingers~2 and~3, so that the contact forces resulting from the mechanism satisfy $f_{21}=f_{22}+f_{23}$. Furthermore, the positive offset $q_x$ indicates that the object lies closer to finger~1 than to fingers~2 and ~3, which is consistent with the compression of $S_{1i}$.} 
\vspace{-0.6cm}
\begin{figure}
    \centering
    \includegraphics[width=0.9\linewidth]{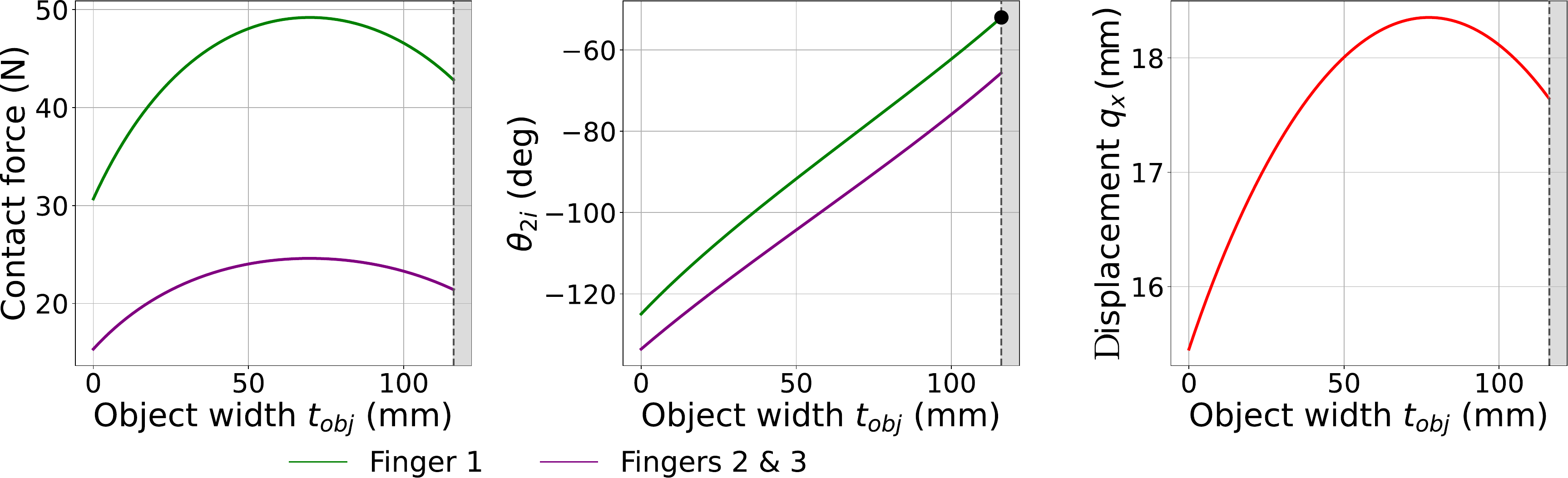}
    \caption{{Overview of cylindrical grasp performances: contact forces, finger flexions ($\theta_{2i}$) and object displacement as a function of the object width.}}
    \label{fig:cyl}
\end{figure}
\vspace{-0.6cm}
{In order to demonstrate the limits of the hand in achieving a stable grasp, the joint limits are investigated (Fig.~\ref{fig:cyl}). The maximum angular limit of $\theta_{21}$ (-52°) is reached for a maximum object width of 116 mm. The minimum limit of $\theta_{2i}$ (-140°) is never reached, indicating that the hand is capable of grasping very thin objects such as a sheet of paper.}
Moreover, the effects of the lever arm and the influence of the finger geometry are evident in both the contact forces and the resulting object displacement, which follow the observed concave trend. {The maximum contact force occurs for an object width of 70mm.} Assuming the local behavior around equilibrium is conservative, the stability of the equilibrium configuration can be assessed by examining a local Hessian of the potential energy with respect to the object displacement along $\mathbf{x}_{hand}$:
\begin{equation}
    H = \frac{d^2U}{dx^2}=-\frac{df_x}{dx}.
    \label{eq:Hessian}
\end{equation}
$f_x$ denotes the sum of the contact forces acting on the object along  $\mathbf{x}_{hand}$.
The positivity of this term indicates a local minimum of the potential energy $U$, meaning that any small displacement increases the potential energy and generates a restoring force to the equilibrium position. Consequently, a small perturbation around $\mathbf{x}_{hand}$ does not compromise stability.

\begin{figure}
    \centering
    \includegraphics[width=0.9\linewidth]{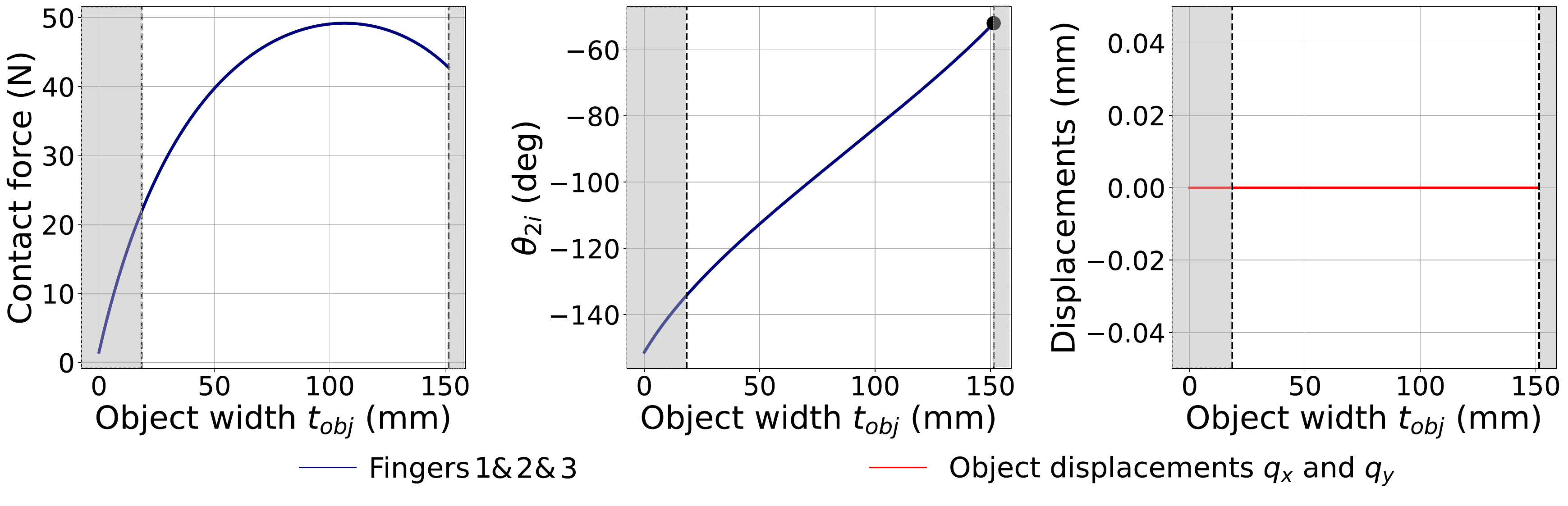}
    \caption{{Overview of spherical grasp performances: contact forces, finger flexions ($\theta_{2i}$) and object displacement as a function of the object width}}
    \label{fig:sph}
\end{figure}

\noindent\textbf{Spherical grasp:}
When the object's center aligns with the hand's center in a spherical grasp, there is no object displacement or spring compression,
as illustrated in Fig.~\ref{fig:sph}.  Then, the three fingers are symmetric which implies $f_{21}=f_{22}=f_{23}$ since $\theta_{11}=\theta_{12}=\theta_{13}=0^\circ$. {As in the cylindrical grasp, the concave trend of the contact force curve confirms the influence of the finger kinematics and the lever-arm effect on the resulting force. For the spherical grasp, possible collisions between fingers impose a minimum admissible object width, given by: $t_{obj,min}=n/\sqrt{3}=18.4\text{mm}$ where $n= 31.9\text{mm}$ the finger width. The maximum width is linked, as for the first scenario, to $l_1$ and the finger range of flexion leading to a maximum object width of 150 mm.}
As above, the goal is to evaluate how small perturbations around $\mathbf{x}_{hand}$ and $\mathbf{y}_{hand}$ influence the stability under a local conservativity assumption. The local Hessian is defined as:
\begin{equation}
H =
\begin{pmatrix}
H_{xx} & H_{xy} \\
H_{yx} & H_{yy}
\end{pmatrix}
=
\begin{pmatrix}
-\dfrac{\partial f_x}{\partial x} & -\dfrac{\partial f_x}{\partial y} \\
-\dfrac{\partial f_y}{\partial x} & -\dfrac{\partial f_y}{\partial y}
\end{pmatrix}.
\label{Hessian2}
\end{equation}
$f_x$ and $f_y$ denote, respectively, the sum of the contact forces acting on the object along $\mathbf{x}_{hand}$ and $\mathbf{y}_{hand}$.
For a stable equilibrium, the eigenvalues must be positive. The results show that around the equilibrium point, $H_{xy}=H_{yx}=0$, meaning that the system is independent in both directions. Moreover, $H_{xx} = H_{yy} >0$, which is consistent with the geometric symmetry of the configuration. Since the stability condition is satisfied, the equilibrium is locally stable.
\section{Conclusions and future work}
This study examined the stability of objects grasped by underactuated robotic hands, focusing on the influence of the mechanism kinematics on stable positions and on the range of admissible object sizes. {The proximal phalanx length, the admissible finger flexion range, as well as the phalanx width have a significant impact on the admissible object range.} Stability analysis and energy-based criteria are used to demonstrate that the system can maintain a stable equilibrium for two canonical grasp configurations (cylindrical and spherical). These insights provide a foundation for optimizing grasp performance.
Future work will expand this analysis in several ways. First, we will incorporate support-object and fingertip frictions, enabling the study of the entire grasping process, from initial contact to lift-off. Additionally, we will explore enveloping grasps and interactions with arbitrarily shaped objects to better characterize hand capabilities. Further steps include determining the maximum graspable mass and optimizing hand design for specific objects. Finally, integrating this approach with learning-based controllers could merge passive stability and active dexterity, enabling versatile and energy-efficient grasping in dynamic environments.

{\vspace{-0.5cm}}
\section*{Acknowledgment}
This work was supported by funding from the French government, managed by the National Research Agency under the France 2030 program, reference ANR-22-EXOD-0003, within the PEPR Organic Robotics.
{\vspace{-0.5cm}}

\end{document}